%% file: beam.tex
\documentclass[prl,twocolumn,amsmath,nobibnotes]{revtex4-1}

\usepackage[utf8]{inputenc}
\usepackage{graphicx, subfig, floatrow}
\usepackage{amssymb,amsfonts,amsmath, dsfont}
\usepackage{multirow}

\usepackage{natbib}

\usepackage[]{algorithm2e}

\usepackage{hyperref}

\DeclareRobustCommand{\Fig}[1]{Figure~\ref{fig:#1}}
\DeclareRobustCommand{\Eq}[1]{Equation~\ref{eq:#1}}

\DeclareRobustCommand{\Tab}[1]{Table~\ref{tab:#1}}
\DeclareRobustCommand{\Algo}[1]{Algorithm~\ref{algo:#1}}
\DeclareRobustCommand{\Algos}[2]{Algorithms~\ref{algo:#1} and \ref{algo:#2}}

\def \g{\gamma}    
    \def \b{\beta}

\def \E{\textrm{E}} 
\def \Var{\textrm{Var}} 
\def \Cov{\textrm{Cov}} 

\def \del{\partial}    
\def \df{\textrm{d}}  

\def \>{\rangle} 
\def \<{\langle}

\def\cL{\mathcal{L}}
\def\cA{\mathcal{A}}
\def\cC{\mathcal{C}}

\def\DKL{D_{\rm KL}}
\def\bv{\textbf{v}}
\def\bh{\textbf{h}}
\def\data{\textrm{data}}
 
\def\be{\begin{equation}} 
\def\ee{\end{equation}} 
\def\longrightharpoonup{\relbar\joinrel\rightharpoonup}
\def\longleftharpoondown{\leftharpoondown\joinrel\relbar}

\def\longrightleftharpoons{
  \mathop{
    \vcenter{
      \hbox{
      \ooalign{
        \raise1pt\hbox{$\longrightharpoonup\joinrel$}\crcr
	  \lower1pt\hbox{$\longleftharpoondown\joinrel$}
	  }
      }
    }
  }
}

\newcommand \bea {\begin{eqnarray}} 
\newcommand \eea {\end{eqnarray}}

\makeatletter
\newcommand*{\defeq}{\mathrel{\vcenter{\baselineskip0.5ex \lineskiplimit0pt
                     \hbox{\scriptsize.}\hbox{\scriptsize.}}}%
                     =}
\makeatother

\begin{document}

\title{Boltzmann Encoded Adversarial Machines}

\author{Charles K. Fisher}
\email{drckf@unlearn.ai}
\thanks{authors listed alphabetically.}
\affiliation{Unlearn.AI, Inc., San Francisco, CA 94108}

\author{Aaron M. Smith}
\affiliation{Unlearn.AI, Inc., San Francisco, CA 94108}

\author{Jonathan R. Walsh}
\affiliation{Unlearn.AI, Inc., San Francisco, CA 94108}

\date{\today}

\begin{abstract} 
Restricted Boltzmann Machines (RBMs) are a class of generative neural network that are typically trained to maximize a log-likelihood objective function.  We argue that likelihood-based training strategies may fail because the objective does not sufficiently penalize models that place a high probability in regions where the training data distribution has low probability.  To overcome this problem, we introduce Boltzmann Encoded Adversarial Machines (BEAMs). A BEAM is an RBM trained against an adversary that uses the hidden layer activations of the RBM to discriminate between the training data and the probability distribution generated by the model.  We present experiments demonstrating that BEAMs outperform RBMs and GANs on multiple benchmarks.
\end{abstract} 

\maketitle

\section{Introduction}

A machine learning model is {\it generative} if it learns to draw new samples from an unknown probability distribution. Generative models have two important applications. First, generative models enable simulations of systems with unknown, or very complicated, mechanistic laws. For example, generative models can be used to design molecular compounds with desired properties~\cite{kadurin2017drugan}. Second, in the process of learning to generate samples from a distribution a generative model must learn a useful representation of the data. Therefore, generative models enable unsupervised learning with unlabeled data~\cite{hinton1999unsupervised}. 

The last decade has produced revolutionary advances in machine learning, largely due to progress in training neural networks. Much of this progress has been on {\it discriminative} models rather than generative models. Still, neural generative models such as Restricted Boltzmann Machines (RBMs)~\cite{hinton2006reducing, salakhutdinov2009deep}, Variational Autoencoders (VAEs)~\cite{kingma2013auto, rolfe2016discrete, kuleshov2017neural}, and Generative Adversarial Networks (GANs)~\cite{goodfellow2014generative} have demonstrated promising results on a number of problems. GANs, in particular, are generally regarded as the current state-of-the-art~\cite{karras2017progressive}.

Unlike most other generative models, GANs are trained to minimize a distance between the data and model distributions rather than to maximize the likelihood of the data under the model~\cite{arjovsky2017towards, nowozin2016f}. As a result of the form of this distance function, and because they are built on feedforward neural networks, typical formulations of GANs can be trained using standard backpropogation \cite{rumelhart1986learning}. However, GANs have their drawbacks. GAN training can be difficult and unstable~\cite{arjovsky2017towards, arjovsky2017wasserstein}. Moreover, although one of the main advantages of GANs is that they can be trained end-to-end using backpropogation, recent state-of-the-art approaches have used a layerwise training strategy~\cite{karras2017progressive} reminiscent of methods used to train Deep Boltzmann Machines~\cite{hinton2012better}.

\begin{figure}[t]
\subfloat[][Generative Adversarial Network]{
\includegraphics[width=3in]{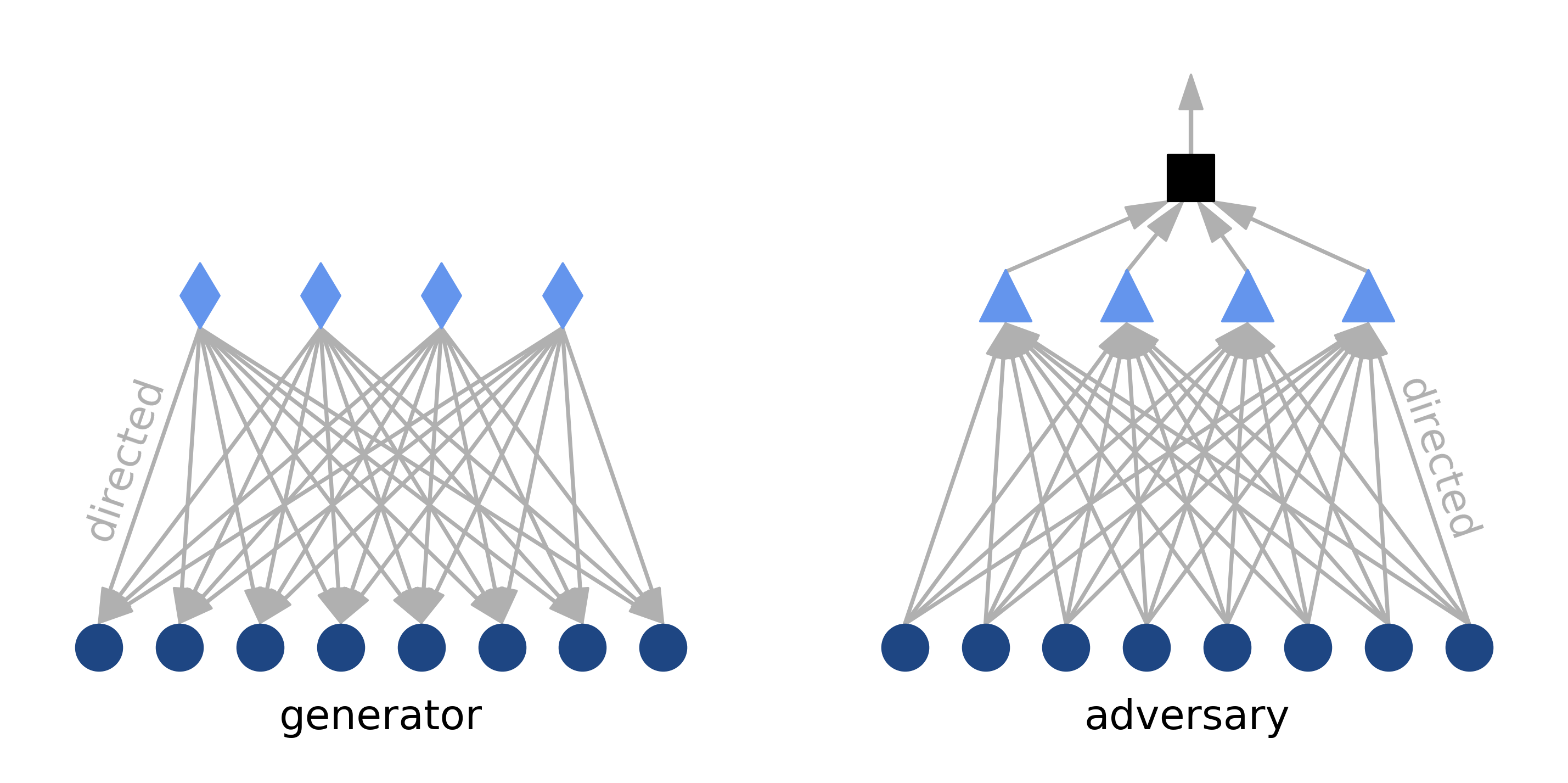}}
\\
\subfloat[][Boltzmann Encoded Adversarial Machine]{
\includegraphics[width=3in]{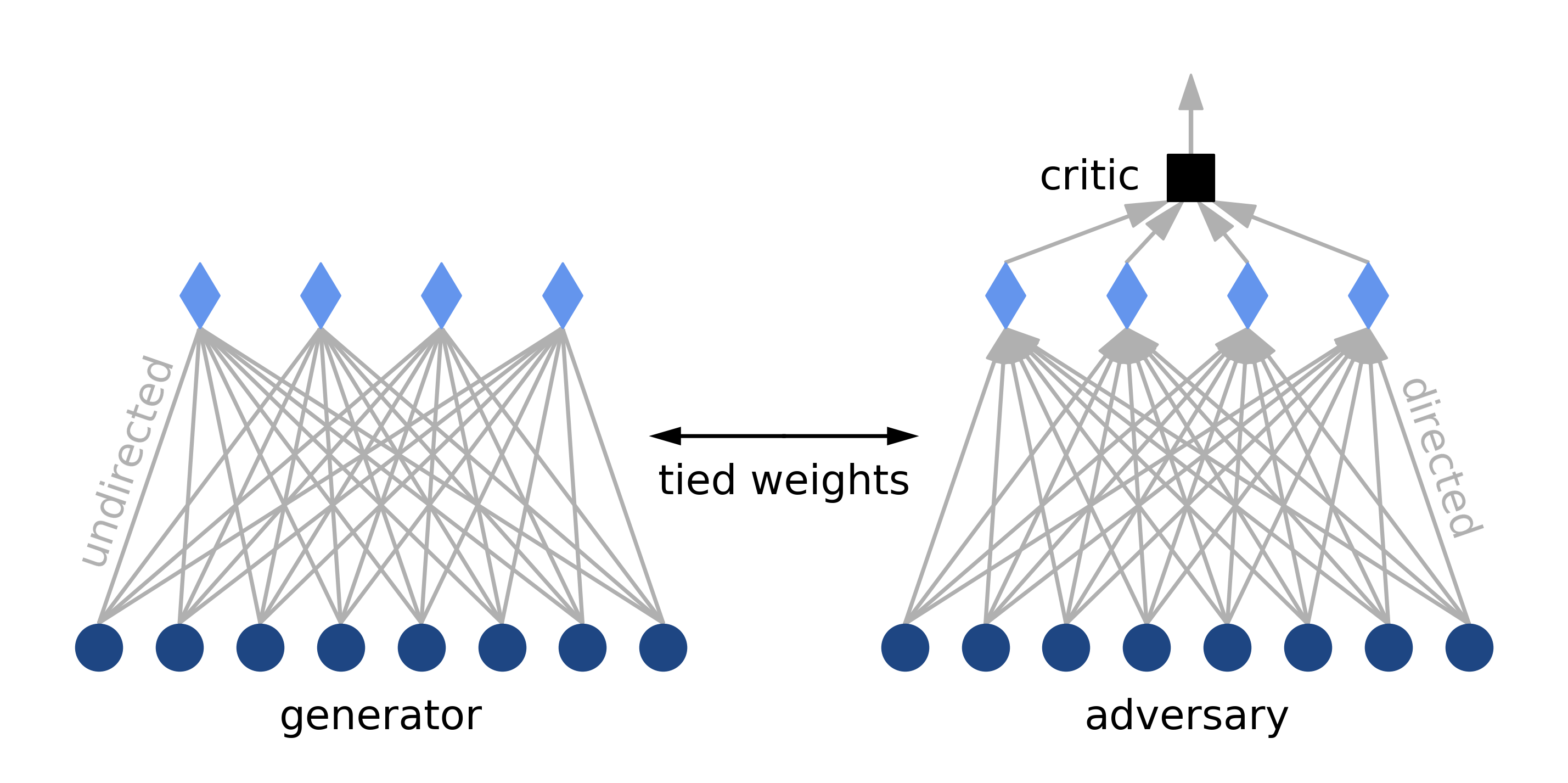}} 
\caption{{\bf Architecture of a BEAM}. 
(a) The generator of a GAN is a feed-forward neural network that transforms random noise into an image, and the adversary is a feed-forward neural network classifies the input image. (b) A BEAM uses an RBM generator trained to minimize an objective function that combines the negative log-likelihood and an adversarial loss. The adversarial loss is computed by a critic trained on the activations of the hidden units of the generator.
\label{fig:beam-architecture}}
\end{figure}

The popularity of RBM-based generative models, including Deep Belief Networks and Deep Boltzmann Machines, has faded in recent years. The charge is that other approaches, especially GANs, simply work better in practice. However, RBM based architectures do have some advantages. For example, RBMs can be easily adapted for use on multimodal data sets~\cite{srivastava2012multimodal} and on time series~\cite{taylor2007modeling, taylor2009factored, sutskever2009recurrent} without major modifications, and RBMs allow one to perform both generation and inference with a single model. Given that RBMs and derived models generally have sufficient representational power to learn essentially any distribution~\cite{le2008representational}, the difficulties must arise during training. 

In this work, we take inspiration from GANs to propose a new method for training RBMs. We call the resulting model a Boltzmann Encoded Adversarial Machine (BEAM; see \Fig{beam-architecture}).  While the adversarial concept used in BEAMs is similar to GANs, there are some distinct features.  The primary one is that the adversary operates on the hidden layer activations of the RBM.  Because the latent variable representation from the RBM is a consolidated representation of the visible units, simple adversaries -- even ones that do not need to be trained -- are often sufficient to obtain good results.  This makes training simple and stable.  Furthermore, we obtain our best results by optimizing a convex combination of the log-likelihood and adversarial objectives.  The component of the objective from the log-likelihood allows the training data to play an active role in determining the gradient (while it only plays a passive role as part of the discriminator in the adversarial gradient, as it also does in GANs).

BEAMs achieve excellent results on a variety of applications, from low dimensional benchmark datasets to higher dimensional applications such as image generation, outperforming GANs of similar or higher complexity. 
These results indicate that BEAMs provide a powerful approach to unsupervised learning.

This paper is structured as follows.  We begin with a brief review of RBMs, then discuss some problems with maximum likelihood training of RBMs and go on to define and describe BEAMs. Finally, we present the results of experiments comparing RBMs, GANs, and BEAMs and discuss.

\begin{figure}[t]
\includegraphics[width=3.25in]{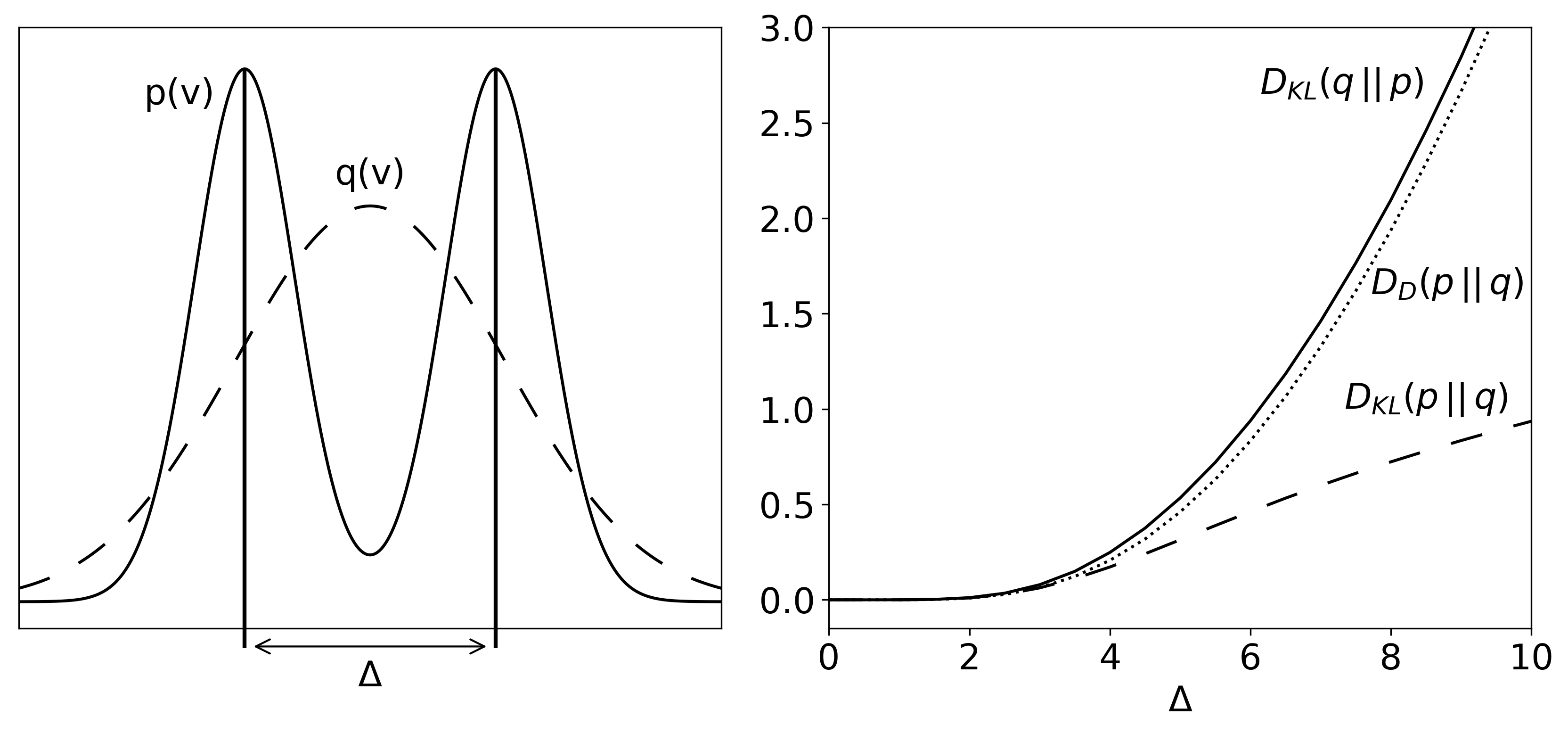}
\caption{{\bf Comparison of distances between distributions}. We consider the distance between $p(v)$, a mixture of two Gaussian distributions separated by a distance $\Delta$, and $q(v)$, a single Gaussian distribution with the same mean and standard deviation as $p(v)$. The forward KL divergence $\DKL(p \,\|\, q)$ increases slowly as $\Delta$ increases, while the reverse KL divergence $\DKL(q \,\|\, p)$ and discriminator divergence $D_D (p \,\|\, q)$ increase rapidly.
\label{fig:kldivergence}}
\end{figure}

\section{Theory and Methods}

\subsection{Restricted Boltzmann Machines
\label{sec:RBM}}

An RBM is an energy based model with two layers of neurons. The visible units $\bv$ describe the data and the hidden units $\bh$ capture interactions between the visible units. The joint probability distribution $p(\bv, \bh) = Z^{-1} e^{-E(\bv, \bh)}$ is defined by an energy function:
\be \label{eq:energy-function}
E(\bv, \bh) = - \sum_i a_i(v_i) - \sum_{\mu} b_{\mu}(h_{\mu}) - \sum_{i \mu} W_{i \mu} \frac{v_i}{\sigma_i^2} \frac{h_{\mu}}{\epsilon_{\mu}^2}
\ee
with a partition function $Z = \int \df \bv \, \df \bh \, \exp\bigl( -E(\bv, \bh) \bigr)$. This formulation, where $a_i(\cdot)$ and $b_{\mu}(\cdot)$ are generic functions and $\sigma_i$ and $\epsilon_i$ are scale parameters, is a flexible way of writing a generic RBM that encompasses common models such as Bernoulli RBMs and Gaussian RBMs. The key feature of an RBM is the conditional independence of the layers, i.e. $p(\bv | \bh) = \prod_i p_i(v_i | \bh)$ and $p(\bh | \bv) = \prod_{\mu} p_{\mu}(h_{\mu} | \bv)$, which allows one to sample from the distribution using block Gibbs sampling.

RBMs are typically trained to maximize the log-likelihood $\cL = \< \log \int \df \bh \, p(\bv, \bh) \>_{data}$ using algorithms such as Persistent Contrastive Divergence (PCD)~\cite{tieleman2008training, hinton2006training}. The derivative of the log-likelihood with respect to a model parameter $\theta$ takes the form~\cite{ackley1985learning}:
\be \label{eq:log-likelihood-derivs}
\del_{\theta} \cL = \< -\del_{\theta} E(\bv, \bh) \>_{data} - \< -\del_{\theta} E(\bv, \bh) \>_{model} \,.
\ee
The two averages are computed using samples from the data set and samples drawn from the model by Gibbs sampling, respectively. We refer the reader to foundational works such as~\cite{hinton2010practical} for more detail.

\subsection{The Problem with Maximum Likelihood
\label{sec:max-like}}

A generative model defined by parameters $\theta$ describes the probability of observing a visible state $\bv$. Therefore, training a generative model involves minimizing a distance between the distribution of the data, $p_{d}(\bv)$, and the distribution defined by the model, $p_{\theta}(\bv)$. The traditional algorithms for training RBMs maximize the log-likelihood, which is equivalent to minimizing the forward Kullback-Liebler (KL) divergence~\cite{kullback1951information}:
\be \label{eq:kl_divergence}
\DKL(p_{d} \,\|\, p_{\theta}) = \int \df \bv \, p_{d}(\bv) \log \bigg( \frac{p_{d}(\bv)}{p_{\theta}(\bv)} \bigg) \,.
\ee
To illustrate some problems with maximum likelihood, we will compare the forward KL divergence to the reverse KL divergence,
\be \label{eq:revkl_divergence}
\DKL(p_{\theta} \,\|\, p_{d}) = \int \df\bv \, p_{\theta}(\bv) \log \bigg( \frac{p_{\theta}(\bv)}{p_{d}(\bv)} \bigg) \,.
\ee

The forward KL divergence, $\DKL(p_{d} \,\|\, p_{\theta})$, accumulates differences between the data and model distributions {\it weighted by the probability under the data distribution}. The reverse KL divergence, $\DKL(p_{\theta} \,\|\, p_{d})$, accumulates differences between the data and model distributions {\it weighted by the probability under the model distribution}. As a result, the forward KL divergence strongly punishes models that {\it underestimate} the probability of the data, whereas the reverse KL divergence strongly punishes models that {\it overestimate} the probability of the data. \Fig{kldivergence} illustrates the difference between the metrics on a simple problem.

There are a variety of sources of stochasticity that enter into the training of an RBM. For example, the model moments have to be estimated using random sampling by Markov Chain Monte Carlo methods, and the gradients are almost always computed from minibatches of data rather than the whole data set. The stochasticity implies that different models may become statistically indistinguishable if the differences in their log-likelihoods are smaller than the errors in estimating them. This creates an {\it entropic force} because there will be many more models with a small $\DKL(p_{d} \,\|\, p_{\theta})$ than there are models with both a small $\DKL(p_{d} \,\|\, p_{\theta})$ and $\DKL(p_{\theta} \,\|\, p_{d})$. As a result, training an RBM using a standard approach with PCD decreases $\DKL(p_{d} \,\|\, p_{\theta})$ (as it should) {\it but tends to increase} $\DKL(p_{\theta} \,\|\, p_{d})$. This leads to distributions with spurious modes and/or to distributions that are oversmoothed. 

\subsection{Advantages of Adversarial Training}
\label{sec:discriminator-div}

One can imagine overcoming the limitations of maximum likelihood training of RBMs by minimizing a combination of the forward and reverse KL divergences. Unfortunately, computing the reverse KL divergence requires knowledge of $p_{d}$, which is unknown. Therefore, we introduce a new type of f-divergence that we call a {\it discriminator divergence}
\be \label{eq:discriminator_divergence}
D_D (p_{d} \,\|\, p_{\theta}) \defeq -\int \df\bv \, p_{\theta}(\bv) \log \bigg( \frac{ 2 p_{d} (\bv)}{p_{d} (\bv) + p_{\theta} (\bv)} \bigg) \,,
\ee
Notice that the optimal discriminator between $p_{d}$ and $p_{\theta}$ will assign a posterior probability
\be
p(\data | \bv) = \frac{p_{d}(\bv)}{p_{d}(\bv) + p_{\theta}(\bv)}
\label{eq:optimal-discriminator}
\ee
that the sample $\bv$ was drawn from the data distribution. Therefore, we can write the discriminator divergence as
\be 
D_D (p_{d} \,\|\, p_{\theta}) = - \log 2 -\int \df\bv \, p_{\theta}(\bv) \log \left(p(\data | \bv) \right)
\ee
to show that it measures the probability that the optimal discriminator will incorrectly classify a sample drawn from the model distribution as coming from the data distribution.

The discriminator divergence belongs to the class of f-divergences defined as $D_{f}(p || q) \defeq \int \df x q(x) f(p(x) / q(x))$. The function that defines the discriminator divergence is
\be
f(t) = \log\left( \frac{t+1}{2t} \right) 
\ee
which is convex with $f(1) = 0$ as required. It is easy to show that the discriminator divergence upper bounds the reverse KL divergence:
\begin{align*}
\log 2 + D_D (p_{d} \,\|\, p_{\theta}) &= \int \df\bv \, p_{\theta}(\bv) \log \left( 1 + \frac{p_{\theta}(\bv)}{p_d(\bv)} \right) \\ \nonumber
&\ge \DKL(p_{\theta} \,\|\, p_{d}) \,. \nonumber
\end{align*}
We introduce this relationship because we usually do not have access to $p_{d}(\bv)$ directly and cannot compute the reverse KL divergence. However, we can train a discriminator to approximate Equation \ref{eq:optimal-discriminator} and, therefore, can approximate the discriminator divergence.

A generator that is able to trick the discriminator so that $p(\data | \bv) \approx 1$ for all samples drawn from $p_{\theta}$ will have a low discriminator divergence. The discriminator divergence closely mirrors the reverse KL divergence and strongly punishes models that {\it overestimate} the probability of the data (\Fig{kldivergence}). Therefore, as with GANs, we hypothesized that it may be possible to improve the training of RBMs using an adversary. Some previous research in this direction includes the Wasserstein RBM~\cite{montavon2016wasserstein} and Associate Adversarial Networks~\cite{arici2016associative}.

\subsection{Boltzmann Encoded Adversarial Machines (BEAMs)
\label{sec:beams}}

We introduce a method -- called a Boltzmann Encoded Adversarial Machine (BEAM) -- for training an RBM against an adversary.  A BEAM minimizes a loss function that is a combination of the negative log-likelihood and an adversarial loss. The adversarial component ensures that BEAM training performs a simultaneous minimization of both the forward and reverse KL divergences, which prevents the oversmoothing problem observed with regular RBMs. 

The architecture of a BEAM is very simple, and is illustrated in \Fig{beam-architecture}.  The RBM (the generative model) is trained with an objective,
\be \label{eq:main-objective}
\cC = -\g \cL - (1-\g) \cA \,,
\ee
that includes a contribution from an adversarial term, $\mathcal{A}$. In theory, the adversary could be any model that can be trained to approximate the optimal discriminator. 

We take inspiration from GANs and train the RBM against a critic function. However, we use a critic function $T(\bh)$ that acts on the hidden unit activations rather than the visible units. That is, the adversary uses same architecture and weights as the RBM, and encodes visible units into hidden unit activations. These hidden unit activations, computed for both the data and fantasy particles sampled from the RBM, are used by a critic to estimate the distance between the data and model distributions. Thus, the BEAM adversarial term is
\begin{align} \label{eq:adversarial_loss}
\cA \defeq \int \df \bh \, p_{\theta} (\bh) T(\bh) \,.
\end{align}
This term has a straightforward interpretation: for any sensible critic, it is minimizing the distance between the marginal distributions of the hidden units under the data and model distributions.

For example, suppose that we had access to the optimal discriminator (on the hidden units):
\be
p(\data | \bh) = \frac{p_{d}(\bh)}{p_{d}(\bh) + p_{\theta}(\bh)}
\ee
where $p_{d}(\bh) \defeq \int d\bv p_{\theta}(\bh | \bv) p_{d}(\bv)$. Then, we could define a critic to minimize the discriminator divergence (on the hidden units) using $\tilde{T}(\bh) = \log 2 + \log(p(\data | \bh))$. In practice, however, we found that we obtain better results with a linear critic:
\begin{equation} \label{eq:optimal_critic}
T(\bh) = 2 \, p(\data | \bh) -1 \, .
\end{equation}
Therefore, all experiments that follow use a linear critic. We use the $2p-1$ form so that the sign of the critic indicates the best guess of the optimal discriminator, but this choice is not important since it only ends up scaling the derivative by a factor of two.

In practice, of course, we don't have access to the optimal discriminator. The usual remedy for GANs is to co-train a neural network to approximate it. In our case, we hypothesized that a simple approximation to the optimal discriminator will be sufficient because are working with the hidden unit activities of the RBM generator rather than the visible units. Therefore, we simply approximate the optimal critic using nearest neighbor methods. In our examples, we simply store the data and fantasy particles from the previous minibatch and use a distance-weighted nearest neighbor approximation. 

A BEAM can be trained using stochastic gradient descent by computing model averages from persistent fantasy particles in the same way as with maximum likelihood training of an RBM.  The derivative of the adversarial term with respect to a model parameter $\theta$ is
\begin{align} \label{eq:adversarial-deriviative}
\del_{\theta} \cA = \Cov_{\theta}[T(\bh), -\del_{\theta} E(\bv, \bh)] \,,
\end{align}
where the covariance is computed with respect to the model distribution $p_{\theta}$. A derivation of this result is presented in the Supplementary Material. It is also possible to define a critic on the visible units directly, or to use some other method other than a nearest neighbor approximation.  We present some comparisons of BEAMs with other critics in the Supplementary Material.

In the context of most formulations of GANs -- which use feed-forward neural networks for both the generator and the discriminator -- one could say that BEAMs use the RBM as both the generator and as a feature extractor for the adversary.  This double-usage allows us to reuse a single set of fantasy particles for multiple steps of the training algorithm. Specifically, we maintain a single set of $M$ persistent fantasy particles that are updated $k$ times per gradient evaluation. The same set of fantasy particles are used to compute the log-likelihood derivative (\Eq{log-likelihood-derivs}) and the adversarial derivative (\Eq{adversarial-deriviative}). Then, these fantasy particles replace the fantasy particles from the previous gradient evaluation in the nearest neighbor estimates of the critic value. Reusing the fantasy particles for each step means that BEAM training has roughly the same computational cost as training an RBM with PCD.

\subsection{Nearest Neighbor Critics
\label{sec:nn-critics}}

Suppose $X = \{ x_1, \dots, x_N \}$ are i.i.d. samples from an unknown probability distribution with pdf $p(x)$ in $\mathbb{R}^n$.  One simple way to estimate $p(x)$ at an arbitrary point $x$ is to make use of a $k$-nearest-neighbor estimate.  Specifically: fix some positive integer $k$ and compute the $k$ nearest neighbors to $x$ in $X$.  Define $d_k$ to be the distance between $x$ and the furthest of the nearest-neighbors.  Then estimate the density $p(x)$ to be the density of the uniform distribution on a ball of radius $d_k$.  That is,
\be \label{eq:nn-density}
p(x) = k \Big(\frac{{\pi^{\frac{n}{2}}}}{\Gamma(\frac{n}{2} + 1)}d_k^n \Big)^{-1} \,.
\ee

Now denote by $p_{\theta}(x)$ and $p_{d}(x)$ the unknown pdfs of the model and data distributions respectively.  Suppose $X = \{ x_1, \dots, x_{2N} \}$ is a collection of i.i.d. samples exactly half of which are drawn from $p_{\theta}$ and half from $p_{d}$.  We can use the same idea to estimate the ratio $\frac{p_{d}}{p_{d} + p_{\theta}}(x)$.  Fix some $k$ and compute the $k$ nearest neighbors in $X$, denoting by $d_k$ the distance to the furthest.  Then we estimate the denominator as in (\ref{eq:nn-density}).  Let $j$ be the number of nearest neighbors which come from $p_{d}$ as opposed to $p_{\theta}$.  The numerator then can be estimated as uniform on the same size ball with only $j/k$ of the density of the denominator.  As a result the desired estimate is simply the ratio $j/k$.

We put this concept in action by defining the \emph{nearest-neighbor  critic}.  Suppose that we have cached a minibatch of samples from the model and a minibatch of samples from the training dataset.  For any new sample $x$ we can compute the $k$-nearest neighbors from the joined minibatches for some fixed $k$ -- we generally use $k=5$ in examples.  Then the nearest-neighbor critic is defined as the function which assigns to $x$ the ratio $j/k$ in which $j$ is the number of nearest neighbors originating from the data minibatch as opposed to the model minibatch.
\be
T_{NN}(x) \defeq 2 \frac{j}{k} - 1 \,.
\ee

The \emph{distance-weighted nearest-neighbor critic} is a generalization which attempts to add some continuity to the nearest-neighbor critic by applying an inverse distance weighting to the ratio count.  Specifically, let $\{d_0, \dots, d_k\}$ be the distances of the $k$-nearest neighbors in $X$ to some $x$, with $\{d_0, \ldots, d_j\}$ the distances for the neighbors originating from the data samples and $\{d_{j+1}, \ldots, d_k\}$ the distances for the neighbors originating from the model samples.  Then distance-weighted nearest-neighbor critic is defined as:
\be
T_{DNN}(x) \defeq 2 \frac{\sum_{i=1}^j \frac{1}{d_i + \epsilon}}{ \sum_{i=1}^k \frac{1}{d_i + \epsilon} } - 1 \,,
\ee
where $\epsilon$ regularizes the inverse distance.

\subsection{Temperature Driven Sampling}

Finally, we use a simple trick to improve the mixing of the RBM while sampling the fantasy particles. We assign each fantasy particle an independently sampled inverse temperature $\beta$ and define the probability as $p(\bv, \bh) = Z^{-1} e^{-\beta E(\bv, \bh)}$. The inverse temperature is drawn from an autoregressive Gamma process~\cite{gourieroux2006autoregressive} with mean $1$, standard deviation $<1$, and autocorrelation $>0$. For applications in this paper, we set the standard deviation to around $0.9$ and the autocorrelation coefficient to $0.9$, though specific values are noted in the Supplementary Material. The intuition behind this algorithm is similar to parallel tempering~\cite{swendsen1986replica, geyer1991markov, desjardins2010tempered, brakel2012training, desjardins2010parallel, desjardins2014deep}. When $\b$ is small, the fantasy particles will be able to explore the space quickly. Setting the mean to $\b = 1$ ensures that the sampled distribution stays {\it close} to the true distribution, while setting the autocorrelation close to $1$ ensures that the inverse temperatures evolve slowly relative to the fantasy particles, which can remain in quasi-equilibrium. Unlike parallel tempering, this \emph{driven sampling} algorithm does not sample from the exact distribution of the RBM. Instead, the driven sampling algorithm samples from a {\it similar} distribution that has fatter tails. However, the driven sampling algorithm adds little computational overhead and generally improves training outcomes. Some additional details and simulations are provided in the Supplementary Material.

\subsection{Using KL Divergences to Monitor Training}

We monitor both the forward and reverse KL divergences during training. Following {\cite{wang2009divergence}, let $\{X_i\}_{i = 1}^{n}$ and $\{Y_i\}_{i = 1}^{m}$ be samples drawn from densities $p$ and $q$. Let $\rho_n(i)$ be the distance from $X_i$ to its nearest neighbor in $\{X_j\}_{j \neq i}$, and $\nu_m(i)$ be the distance from $X_i$ to its nearest neighbor in $\{Y_i\}$. Then, 
\be
D_{KL}(p||q) \approx \frac{d}{n} \sum_{i = 1}^n \frac{\nu_m(i)}{\rho_n(i)} + \log \frac{m}{n-1}
\ee
where $d$ is the dimension of the space (i.e., the number of visible units). The reverse KL divergence can be computed by reversing the identities of $X$ and $Y$. In practice, we monitor the KL divergences using a held-out validation set consisting of 10\% of the data. For computational reasons, we compute the KL divergences on minibatches of the validation set and then average the values.

\section{Results
\label{sec:results}}

\begin{figure*}[t!]
\includegraphics[width=6.5in]{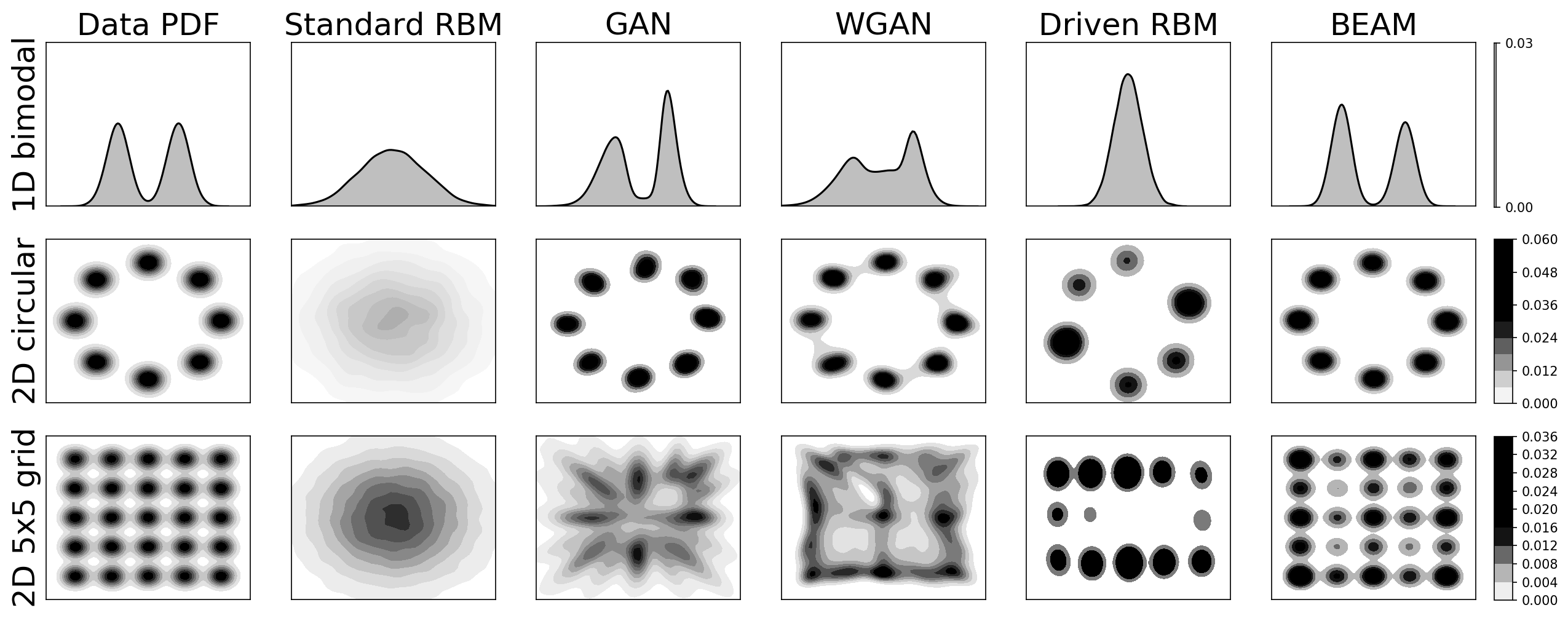}
\caption{{\bf Comparison of generative models on mixtures of Gaussians.} Three datasets constructed from mixtures of Gaussians: a 1-D mixture of two Gaussians, a 2-D mixture of eight Gaussians arranged in a circle, and a 2-D mixture of Gaussians arranged on a 5x5 grid. Distributions of fantasy particles from a standard RBM, a vanilla GAN, and a Wasserstein GAN (WGAN) are compared to distributions of fantasy particles from a RBM trained with a driven sampler and to a BEAM.
\label{fig:mixture-models}}
\end{figure*}

We present empirical results on BEAMs using some datasets that are commonly used to test generative models. We aim to demonstrate three key results: 
\begin{enumerate}
\item RBMs produce poor results because the reverse KL divergence increases during training even though the forward KL divergence decreases. 
\item BEAMs trained with a driven sampler minimize both the forward and reverse KL divergences, leading to better results than RBMs trained by standard methods. 
\item BEAMs produce results that are comparable to, or better than, GANs on multiple benchmarks. 
\item The simplicity of the adversary ensures that BEAM training is stable.
\end{enumerate}

\subsection{Mixture Models}

Our first set of experiments are on a series of 1 and 2-dimensional Mixtures of Gaussians (MoGs) similar to those used in the Wasserstein GAN paper~\cite{pmlr-v70-arjovsky17a}. We compare the results from five different generative models. Models from the literature include a vanilla GAN~\cite{goodfellow2014generative, mnist-gan}, a Wasserstein GAN~\cite{arjovsky2017wasserstein, pmlr-v70-arjovsky17a, WGAN-github}, and a Gaussian-Bernoulli RBM~\cite{cho2013gaussian}. Our models include a Gaussian-Bernoulli RBM trained with the driven sampler and a Gaussian-Bernoulli BEAM with equally weighted likelihood and adversarial losses. All of the RBM based models have the exact same architecture. Details on the model architectures and training parameters are given in the Supplementary Material.

\Fig{mixture-models} shows a comparison of fantasy particles from each of the generative models along with the corresponding data distributions. A standard RBM trained using persistent contrastive divergence with 100 update steps per gradient evaluation fails to learn that the data distribution has multiple modes. Instead, it spreads the model density across the support of the data distribution. The vanilla GAN and the WGAN are both able to learn the 1-D mixture of two Gaussians and the 2-D mixture of eight Gaussians, but fail on the 2-D MoGs arranged in a 5x5 grid. Surprisingly, our results with the vanilla GAN are significantly better than those reported in the literature~\cite{pmlr-v70-arjovsky17a} and are comparable in quality to the results with WGAN. Training the Gaussian-Bernoulli RBM using the driven sampler leads to improvements over the standard RBM. Notably, the BEAM is the only model that learns all three datasets. 

Training an RBM as a BEAM decreases both the forward and reverse KL divergences, as shown in the left panel of \Fig{training-metrics} for the MoGs arranged in a 5x5 grid. In the early stages of training, the BEAM fantasy particles are spread out across the support of the data distribution -- capturing the modes near the edge of the grid. These early epochs resemble the distributions obtained with GANs, which also concentrate density in the modes near the edge of the grid. As training progresses, the BEAM progressively learns to capture the modes near the center of the grid.  

\begin{figure*}[t]
\includegraphics[width=6.0in]{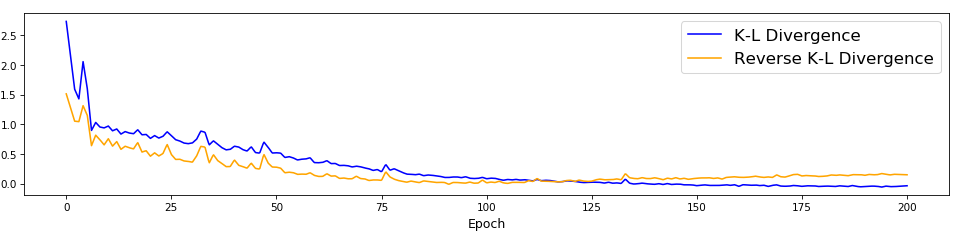} \\
\includegraphics[width=6.0in]{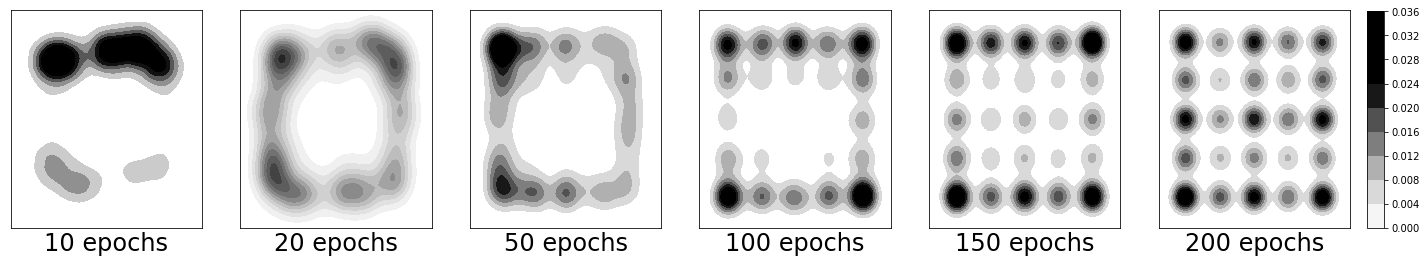}
\caption{{\bf Training a BEAM on a 2-D Mixture of Gaussians (MoGs) arranged in a 5x5 grid.} Top panel shows estimates of the forward KL divergence, $\DKL(p_{d} \,\|\, p_{\theta})$, and the reverse KL divergence, $\DKL(p_{\theta} \,\|\, p_{d})$, per training epoch. Right panels show distributions of fantasy particles at various epochs during training. 
\label{fig:training-metrics}}
\end{figure*}

\subsection{MNIST}

The MNIST dataset of handwritten images~\cite{lecun-mnisthandwrittendigit-2010} is one of the most widely used benchmarks in machine learning. We present results on MNIST with continuous, grayscale images, and MNIST with binary, black and white images.  

We compare five different models on continuous MNIST. A non-convolutional (i.e., fully connected) GAN, a non-convolutional (i.e., fully connected) WGAN, a Gaussian-Bernoulli RBM, a Gaussian-Bernoulli RBM with a temperature driven sampler, and a Gaussian-Bernoulli BEAM. Details of the architectures and training parameters are given in the Supplementary Material. It is important to note that none of these models is designed to produce state-of-the-art results on MNIST; for example, you can get better results using convolutional, rather than fully-connected, networks (see Supplementary Material). However, restricting the analyses to the chosen architectures provides a cleaner comparison of the different training approaches. 

The critic in a BEAM uses the hidden unit activities as features, but these features are not useful during the early stages of training when there is little mutual information between the visible and hidden units of the generator. Therefore, we train the BEAM through two phases. For the first 25 epochs, we use regular maximum likelihood training with persistent contrastive divergence and driven sampling. After 25 epochs, we change the relative weights of the likelihood and the adversary in the loss function to $\g = 0.1$ so that the adversarial term dominates the gradient and train for an additional 30 epochs. The training dynamics are shown in \Fig{mnist-training-metrics}.

RBM based architectures trained by maximum likelihood will decrease the forward KL divergence. This is shown clearly in \Fig{mnist-training-metrics} -- the forward KL divergence decreases during training of the Gaussian-Bernoulli RBM, the Gaussian-Bernoulli RBM with a driven sampler, and the Gaussian-Bernoulli BEAM. However, the figure also clearly shows that the reverse KL divergence increases during training. The training metrics for the BEAM rapidly diverge from the RBMs once the adversary is turned on after epoch 25. The reverse KL divergence of the BEAM quickly drops towards zero while the reverse KL divergence of the RBMs continue to rise. By the end of training, the BEAM obtains comparable, or better, metrics than all other architectures. 

\begin{figure*}[t!]
\includegraphics[width=6.0in]{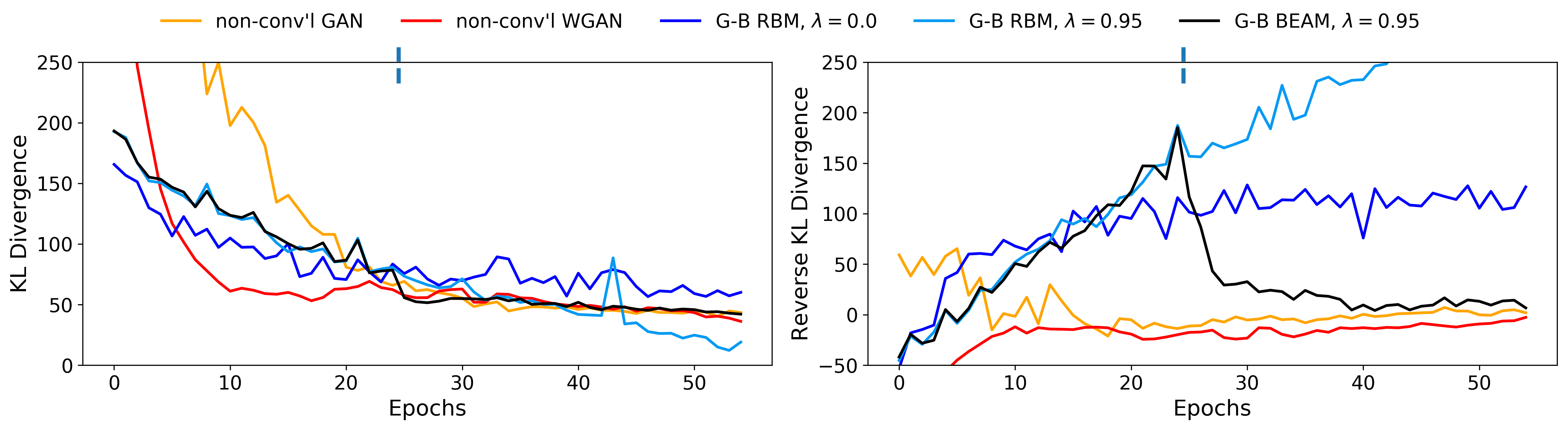}
\caption{{\bf Training metrics on continous MNIST.} The forward KL divergence, $\DKL(p_{d} \,\|\, p_{\theta})$, and the reverse KL divergence, $\DKL(p_{\theta} \,\|\, p_{d})$ divergence during training on MNIST. Both divergences were estimated using an approach based on $k$-nearest neighbors~\cite{wang2009divergence}. Adversarial training for the BEAM begins after epoch 25 (vertical dashed line).
\label{fig:mnist-training-metrics}}
\end{figure*}

Fantasy particles for continuous MNIST are shown in the top row of \Fig{mnist-comparison} along with a table of KL divergences at the end of training. The non-convolutional GAN, non-convolutional WGAN, and the BEAM have similar metrics at the end of training. The errors that they make, however, are qualitatively different. The GANs produce sharp images that are a bit blotchy, whereas the BEAM produces smooth images that are a bit blurry. The regular Gaussian-Bernoulli RBM fails to produce reasonable digits at all, whereas the Gaussian-Bernoulli RBM trained with the driven sampler is a bit better.

\begin{figure*}[t!]
\includegraphics[width=6.5in]{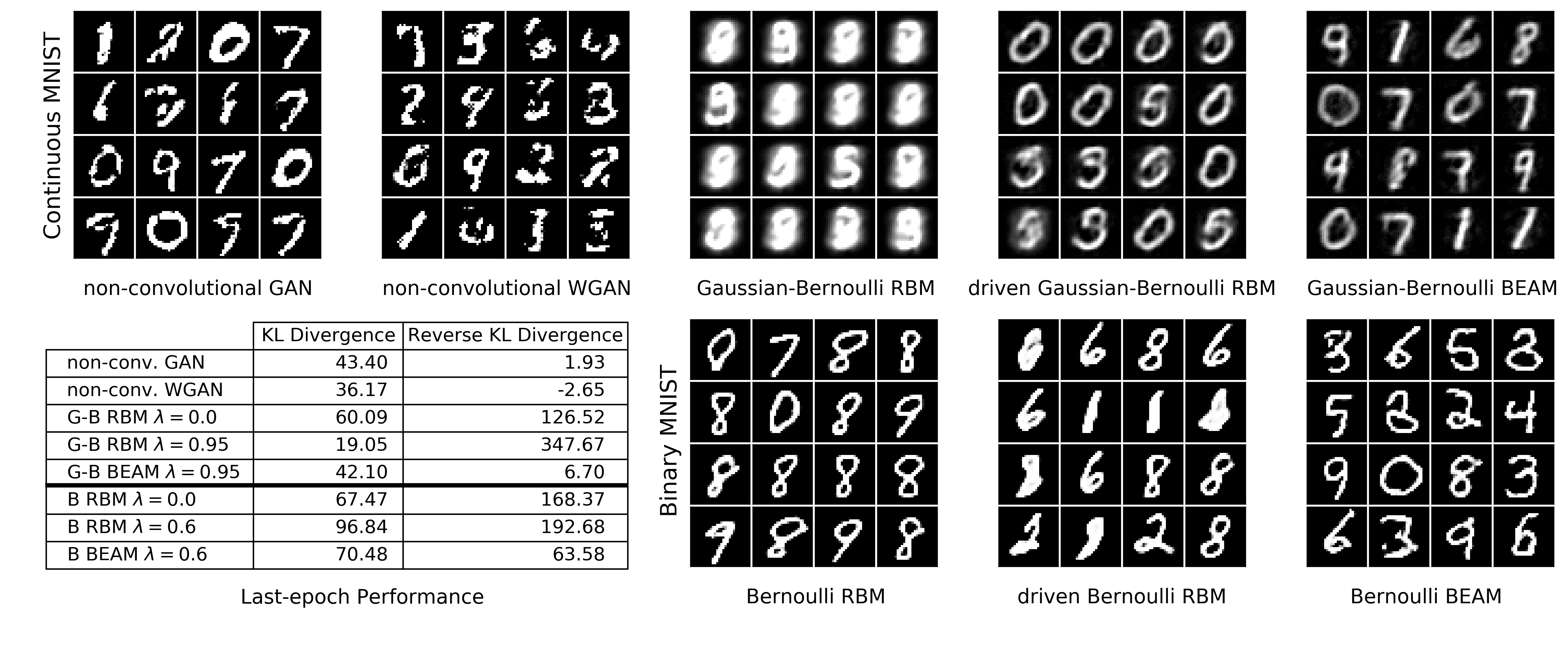}
\caption{{\bf Comparison of MNIST fantasy particles.}  Sixteen particles sampled at random from each of the generators.  The RBM fantasy particles were randomly initialized and sampled for 100 MCMC steps; the figure shows $\< \bv \>_{p_{\theta}(\bv | \bh)}$ computed from the last iteration. Note the thick line in the table of KL divergences separating results on continuous MNIST from those on binary MNIST indicating that these values are not comparable.
\label{fig:mnist-comparison}}
\end{figure*}

A regular Bernoulli-Bernoulli RBM performs a lot better on binary MNIST than a Gaussian-Bernoulli RBM does on continuous MNIST, as shown in the second row of \Fig{mnist-comparison}. Even though a Bernoulli-Bernoulli RBM is well-suited to modeling binary MNIST, it still learns a model distribution with a low forward KL divergence and a high reverse KL divergence, as shown in the table. Adversarial training of the genenerator as a Bernoulli-Bernoulli BEAM fixes this problem leading to a better model.

We do not show any GANs for the binary MNIST problem. In general, it is difficult to train GANs on discrete data due to the inability to backprop through a discrete variable (though, there are ways around this problem as in \cite{yu2017seqgan}). Thus, one advantage of a BEAM is that it is much easier to train on discrete data than a GAN and much easier to train on continous data than a standard RBM.

Throughout, we have presented BEAMs as an adversarial approach to training RBMs where the hidden unit activities of the RBM are used as features for the critic. We claim that this allows us to use a simple classifier to approximate the optimal critic. However, it is possible to train an RBM against an adversary that uses the visible units directly (as in a GAN). Empirically, we have found that applying the critic to the hidden unit activities works better; see Figure S2 for an example.

\subsection{Celebrity Faces}

\begin{figure*}[t]
\subfloat[][BEAM + Decoder]{
\includegraphics[width=1.55in]{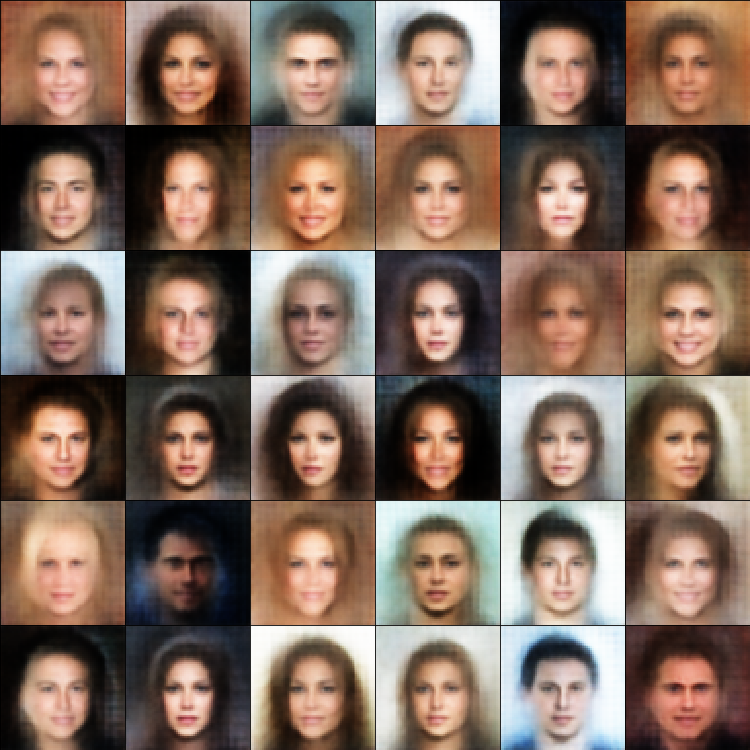}}
\hspace{.1em}
\subfloat[][DCGAN]{
\includegraphics[width=1.55in]{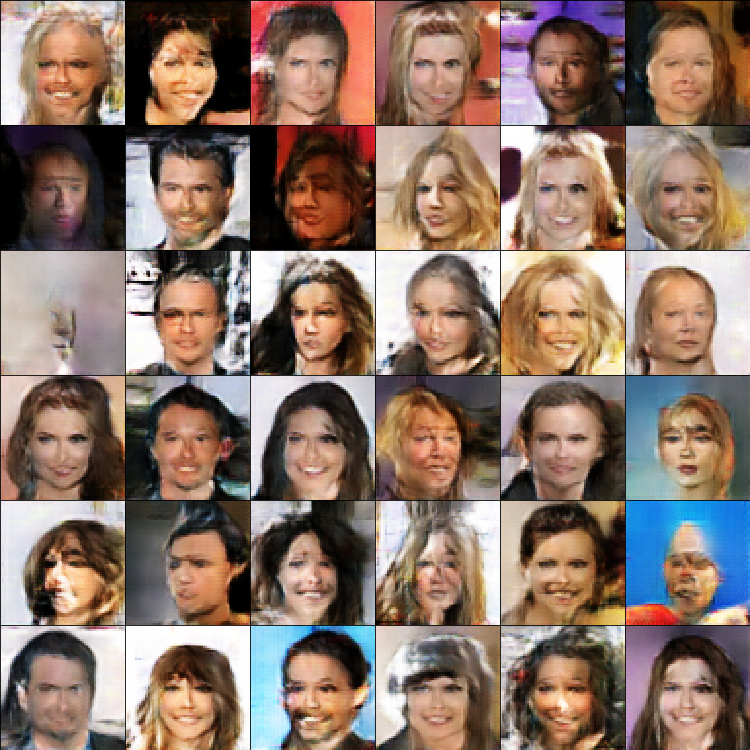}}
\hspace{.1em}
\subfloat[][DCWGAN]{
\includegraphics[width=1.55in]{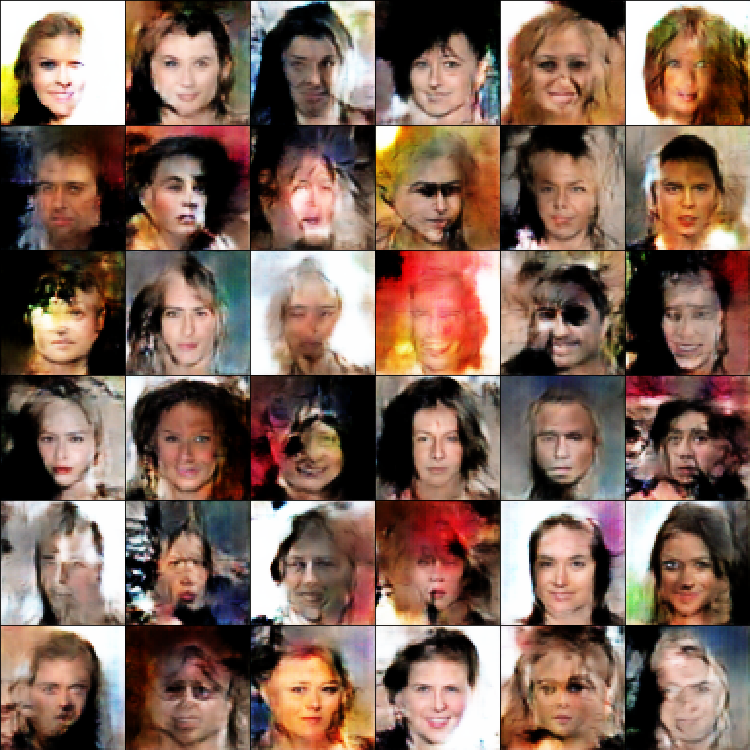}}
\hspace{.1em}
\subfloat[][DCWGAN \cite{li2017mmd}]{
\includegraphics[width=1.55in]{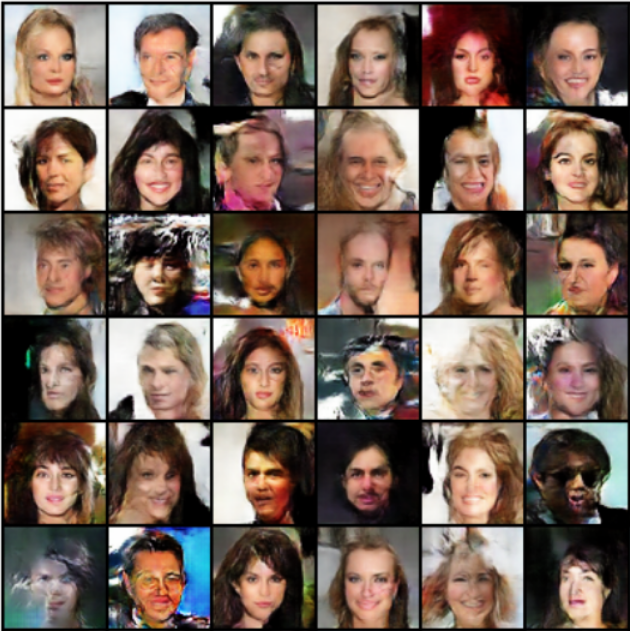}}
\caption{{\bf Comparison of CelebA fantasy particles.} 36 fantasy images sampled from the (a) BEAM with the decoder network, (b) a deep convolutional GAN, and (c) a deep convolutional WGAN. Our implementations were chosen so that each model has very similar architectures. These architectures and training parameters are provided in the Supplementar Material. For comparision, we directly reproduce the CelebA results of a deep convolutional WGAN as reported in \cite{li2017mmd}.
\label{fig:celeb_fantasy}}
\end{figure*}

Natural images present a more complex dataset for which model performance can be easily determined.  We use the CelebA dataset, consisting of $64\times 64 \times 3$ pictures of celebrities' faces, to demonstrate that BEAMs scale to more complex problems. This dataset requires convolutional architectures to obtain good performance.  Because exploring convolutional RBMs is orthogonal to the purpose of this work, we use a separately trained convolutional autoencoder to extract features from the images. These features are used as the visible input to the BEAM.

The autoencoder is trained with sufficient depth and number of features to provide high-quality reconstructions of the data.  This architecture forms the basis of the convolutions used in training the BEAM, DCGAN, and DCWGAN, as shown in Figure S6 in the Supplementary Material. Sample images from the dataset and their reconstructions are shown in Figure S7 in the Supplementary Material.

As with the MNIST examples, we train the CelebA BEAM in two phases.  For the first 15 epochs, we use the log-likelihood objective function with persistent contrastive divergence and driven sampling.  After this phase we train for an additional 50 epochs using the combined log-likelihood-adversarial objective function in \Eq{main-objective} with $\gamma = 0.1$.

For comparison, we train a DCGAN and DCWGAN using the same convolutional architecture as the autoencoder that was used to extract features for BEAM training.  That is, the DCGAN/DCWGAN generator uses an initial fully connected layer followed by the same architecture as the decoder of the autoencoder and the DCGAN/DCWGAN critic uses the same architecture as the encoder of the autoencoder followed by a fully connected layer to a single unit. The DCGAN/DCWGAN share many of architectural features with the autoencoder, but do not share any parameters. Instead, the DCGAN and the DCWGAN were trained end-to-end on CelebA.

Images are generated from the BEAM by sampling fantasy particles and passing them through the decoder.  Example generated images from the BEAM, DCGAN, and DCWGAN are shown in \Fig{celeb_fantasy}.  It is clear that the BEAM images are internally consistent with clear features across each face. However, the images are a bit blurry -- especially towards the corners of the image in the backgrounds. The images produced by the DCGAN and DCWGAN have sharper local features, but notably poorer global correlations. Although the images produced by the GANs are not particuarly high-quality, they are qualitatively similar to results appearing in the literature. To illustrate this, we have directly reproduced fantasy images from a DCWGAN that were published in \cite{li2017mmd} (see \Fig{celeb_fantasy}d).  

We note that it is possible to obtain sharper features from the BEAM at the expense of less consistent images by optimizing training to lower the forward KL divergence at the expense of an increased reverse KL divergence.  Furthermore, additional approaches such as centering layers and using deep models produce notably better images; see Figure S8 in the Supplementary Material for an example using centered layers.

\section{Discussion}

We have introduced a novel formulation of RBMs, called BEAMs, that utilize an adversary acting on the hidden unit activations from the RBM to supplement the traditional likelihood-based training.  The additional adversarial loss term ensures that training minimizes both the forward and reverse KL divergences, allowing the model to accurately learn distributions with multiple modes. We have shown that BEAMs excel at a variety of applications, outperforming GANs that use significantly larger computational budgets.  

As the machine learning community increasingly turns its attention to unsupervised learning problems, it is valuable to place this work into a larger context.  The deep learning revolution has driven tremendous advances on supervised learning problems, and a primary outcome is that feed-forward neural networks have become a powerful tool.  GANs and variational autoencoders can be thought of as a natural extension of the broad learning capacity of neural networks and the flexibility of backpropagation, and are tools of choice in many applications.  This is further supported by the software ecosystem for machine learning, which makes many sophisticated tools easily accessible.

RBMs played an active role in kicking off the deep learning revolution~\cite{hinton2006reducing}, but their development slowed with the increased focus on supervised learning and a general attitude that they were unsuited to more complex problems.  However, there are reasons to believe that RBMs will be fundamental in advancing unsupervised learning:
\begin{itemize}
\item Novel training algorithms and novel model architectures can dramatically improve performance.
\item RBMs have several analytic handles to understand models and develop training strategies.
\item Increased capacity to handle complex datasets can be developed through a progressive set of challenging applications.
\end{itemize}
We hope this work reinforces the promise of RBMs.

\bibliography{beam} 

\input{./supporting/supp}
\end{document}

%% file: supporting/supp.tex
\onecolumngrid
\section{Supplmentary Material}

\subsection{Gradients of the Adversarial Loss
\label{app:adversarial_loss}}

In a general adversarial approach to learning, we train a Boltzmann machine, $p_{\theta}(\bv, \bh)$, to minimize a compound objective function $\cC = -\g \cL + (1-\g) \cA$. The compound objective function represents a tradeoff between maximum likelihood learning ($\g = 1$) and adversarial learning ($\g = 0$). Just as with maximum likelihood, the compound objective function can be optimized using stochastic gradient descent. Using a compound objective function helps to mitigate some of the instability problems that plague traditional GANs. For example, the gradient does not vanish even if the discriminator is completely untrained because there is always the term from the likelihood.  

We need to compute the derivatives of the compound objective function in order to minimize it. The differential operator is linear, so we can distribute it across the two terms:
\be 
\del_{\theta} \cC = \g \del_{\theta} \cL - (1-\g) \del_{\theta} \cA \,.
\ee
The first term can be computed from Equation 2 (Main Text). So, all we need to do is compute the second term. It turns out that derivatives of this form can be computed using a simple formula when the model is a Boltzmann machine. 

Let $T(\bv, \bh)$ be a critic function and
\begin{align}
\mathcal{A} \defeq \int \df\bv \, \df\bh \, p(\bv,\bh) T(\bv,\bh)
\end{align}
be the associated adversarial loss. This formulation reduces to the adversarial loss for a BEAM when $T(\bv, \bh) = T(\bh)$ is independent of the visible units, but we derive the general case. We need to compute $\del_{\theta} \E_{p_{\theta}(\bv, \bh)}[T(\bv, \bh)]$. First, we use the stochastic derivative trick:
\begin{align}
\del_{\theta} \cA
&= \frac{\del}{\del \theta} \int \df\bv \, \df\bh \, p(\bv,\bh) T(\bv,\bh) \nonumber \\
&= \int \df\bv \, \df\bh \, T(\bv,\bh) \frac{\del}{\del \theta} \, p(\bv,\bh) \nonumber \\
&= \int \df\bv \, \df\bh \, T(\bv,\bh) \frac{p(\bv,\bh)}{p(\bv,\bh)} \frac{\del}{\del \theta} \, p(\bv,\bh) \nonumber \\
&= \int \df\bv \, \df\bh \, T(\bv,\bh) p(\bv,\bh)\del_{\theta} \log p(\bv,\bh) \nonumber \\
&= \< T(\bv,\bh) \del_{\theta} \log p(\bv,\bh) \>_{p(\bv,\bh)}
\end{align}
We can write the model distribution of a Boltzmann machine as $p_{\theta}(\bv, \bh) = Z_{\theta}^{-1} e^{-E_{\theta}(\bv, \bh)}$ so that $\log p_{\theta}(\bv, \bh) = -E_{\theta}(\bv, \bh) - \log Z_{\theta}$, with $Z_{\theta} = \int \df\bv \, \df\bh \, e^{-E_{\theta}(\bv, \bh)}$. Then, we have $
\del_{\theta} \log p_{\theta}(\bv, \bh) = -\<-\del_{\theta} E_{\theta}(\bv, \bh)\>_{p_{\theta}(\bv, \bh)} - \del_{\theta} E_{\theta}(\bv, \bh)$. Plugging this in, we find:
\begin{align}
\del_{\theta} \cA
&= -\< T(\bv, \bh)\>_{p_{\theta}(\bv, \bh)} \<-\del_{\theta} E_{\theta}(\bv, \bh)\>_{p_{\theta}(\bv, \bh)} + \< T(\bv, \bh) (-\del_{\theta} E_{\theta}(\bv, \bh)) \>_{p_{\theta}(\bv, \bh)} \\
&= \Cov_{p_{\theta}(\bv, \bh)}[ T(\bv, \bh), -\del_{\theta} E_{\theta}(\bv, \bh)] \,.
\end{align}

\subsection{Details of Temperature Driven Sampling}

\begin{figure}[t]
\includegraphics[width=6in]{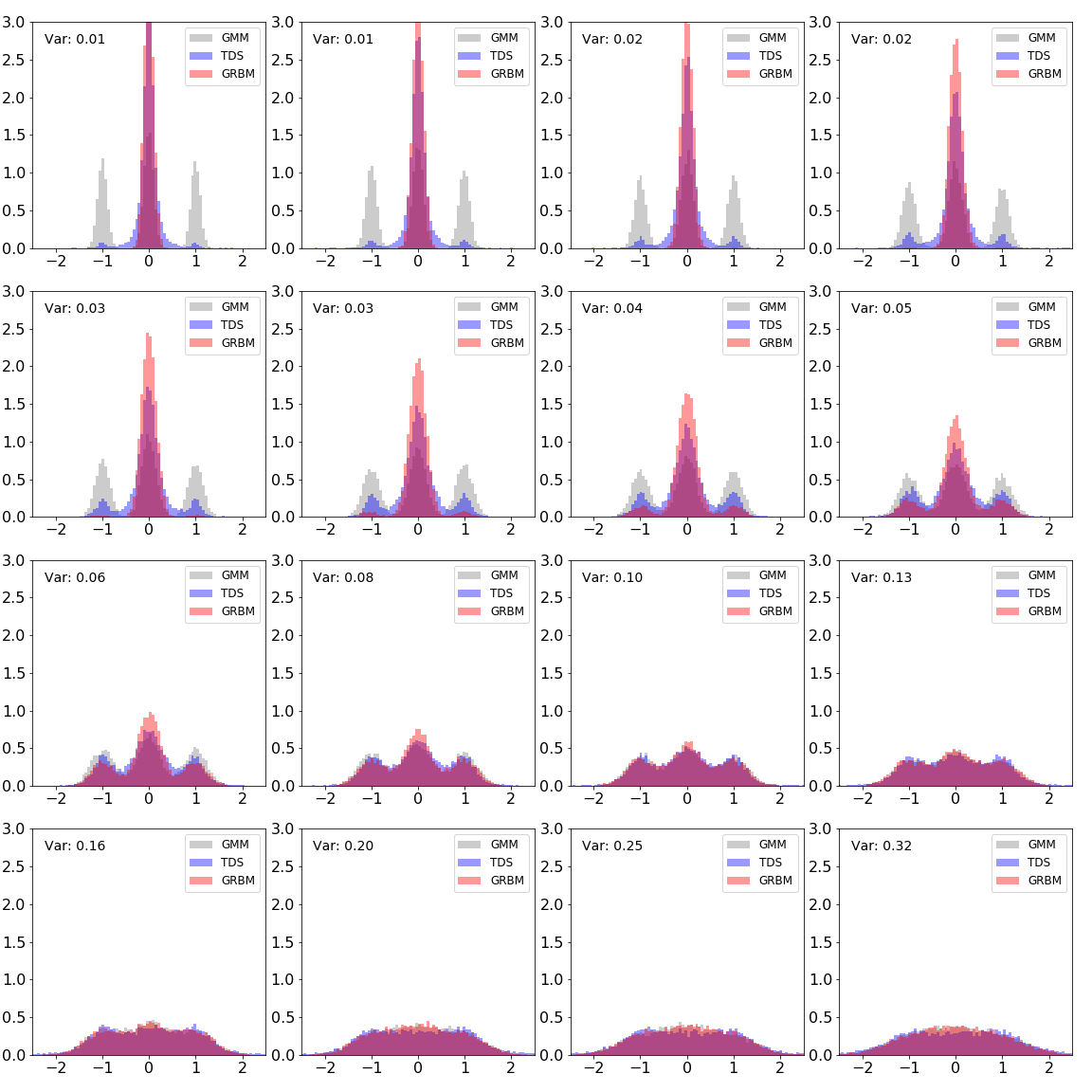}
\caption{{\bf Sampling with a driven sampler.} Comparison of temperature driven sampling (TDS) to regular Gibbs sampling. The RBMs have a single Gaussian visible layer and a softmax hidden layer with 3 hidden units that encode the modes of a mixture of 3 Gaussians. The standard deviation of the inverse temperature was set to $0.9$ for the driven sampler. 
\label{fig:driven_sampler}}
\end{figure}

Our approach to accelerated sampling, which we call Temperature Driven Sampling (TDS), greatly improves the ability to train Boltzmann machines without incurring significant additional computational cost. The algorithm is a variant of a sequential Monte Carlo sampler. A collection of $m$ samples are evolved independently using Gibbs sampling updates from the model. Note that this is not the same as running multiple chains for a parallel tempering algorithm because each of the $m$ samples in the sequential Monte Carlo sampler will be used compute statistics, as opposed to just the samples from the $\b = 1$ chain during parallel tempering. Each of these samples has an inverse temperature that is drawn from a distribution with mean $\E[\b] = 1$ and variance $\Var[\b] < 1$. The inverse temperatures of each sample are independently updated once for every Gibbs sampling iteration of the model; however, the updates are autocorrelated across time so that the inverse temperatures are slowly varying. As a result, the collection of samples are drawn from a distribution that is \emph{close to} the model distribution, but with fatter tails. This allows for much faster mixing, while ensuring that the model averages (computed over the collection of $m$ samples) remain close approximations to averages computed from the model with $\b = 1$. 

\begin{center}
\SetInd{1em}{1em}
\begin{algorithm}[H]
\KwIn{\\
Autocorrelation coefficient $0 \le \phi < 1$.\\
Variance of the distribution $\Var[\b] < 1$. \\
Current value of $\b$.}
\textbf{Set:} $\nu = 1 / \Var[\b]$ and $c = (1-\phi) \Var[\b]$.\\
Draw $z \sim \text{Poisson}(\b * \phi / c)$.\\
Draw $\b' \sim \text{Gamma}(\nu + z, c)$.\\
\Return{$\b'$}
\caption{Sampling from an autocorrelated Gamma distribution.\label{algo:driven-gamma-sampler}}
\end{algorithm}
\end{center}

Details of the TDS algorithm are provided in \Algos{driven-gamma-sampler}{temperature-driven-sampler}. Note that this algorithm includes a standard Gibbs sampling based sequential Monte Carlo sampler in the limit that $\Var[\b] \rightarrow 0$. The samples drawn with the TDS algorithm are \emph{not} samples from the equilibrium distribution of the Boltzmann machine. In principle, it is possible to reweight these samples to correct for the bias due to the varying temperature. In practice, we have not found that reweighting is necessary.  An example of temperature driven sampling applied to a 3-mode MoG is show in \Fig{driven_sampler}.

\begin{center}
\SetInd{1em}{1em}
\begin{algorithm}[H]
\KwIn{\\
Number of samples $m$.\\
Number of update steps $k$.\\
Autocorrelation coefficient for the inverse temperature $0 \le \phi < 1$.\\
Variance of the inverse temperature $\Var[\b] < 1$. \\
}
\textbf{Initialize:}\\
Randomly initialize $m$ samples $\{(\bv_i, \bh_i)\}_{i = 1}^m$.\\
Randomly initialize $m$ inverse temperatures $\b_i \sim \text{Gamma}(1 / \Var[\b], \Var[\b])$.\\
\For{t = 1, \ldots, k}{
\For{i = 1, \ldots, m}{
Update $\b_i$ using \Algo{driven-gamma-sampler}.\\ 
Update $(\bv_i, \bh_i)$ using Gibbs sampling.
}
}
\caption{Temperature Driven Sampling.\label{algo:temperature-driven-sampler}}
\end{algorithm}
\end{center}

\subsection{Details on Models and Training}

\subsubsection{Gaussian Mixtures}
\Tab{gmm_params_table} lays out the parameters of the Gaussian mixture comparison examples.  It is interesting to note just how few parameters are required by the BEAM to model this data.

\subsubsection{MNIST}

\begin{figure*}[t!]
\includegraphics[width=6.0in]{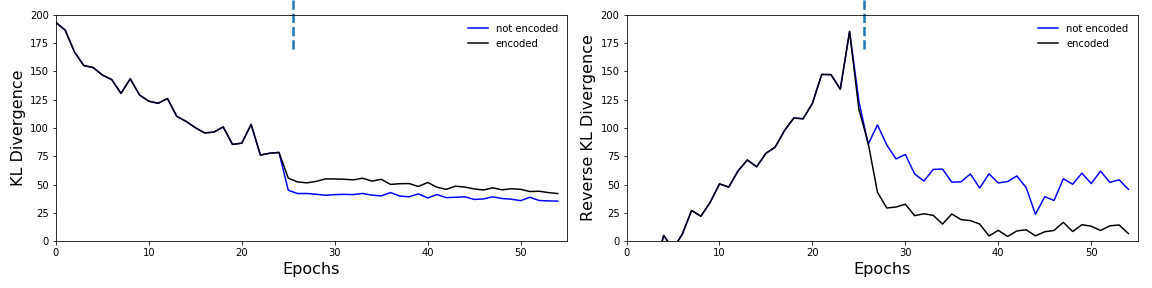}
\caption{{\bf Comparing a BEAM with the critic on the hidden layer to one with the critic on the visible layer.}  The KL divergences of two BEAMs trained on MNIST differing only in whether or not the critic acts on the encoded data or directly on the visible data.  Adversarial training begins after 25 epochs.
\label{fig:mnist-critic-comparison}}
\end{figure*}

\Fig{mnist-critic-comparison} provides a comparison of the training metrics when the discriminator is trained on the hidden unit activities to when the discriminator is trained on the visible units. Both architectures use the same nearest neighbor classifier as the rest of our examples. The two training curves overlay exactly for the first 25 epochs while the generator is pre-trained with maximum likelihood learning. Once the discriminator is turned on, the reverse KL divergence decreases, but training the adversary on the hidden units decreases these metrics much more rapidly.  Table \ref{tab:mnist_params} lays out the parameters of the models.

\begin{figure*}[t!]
\includegraphics[width=6.0in]{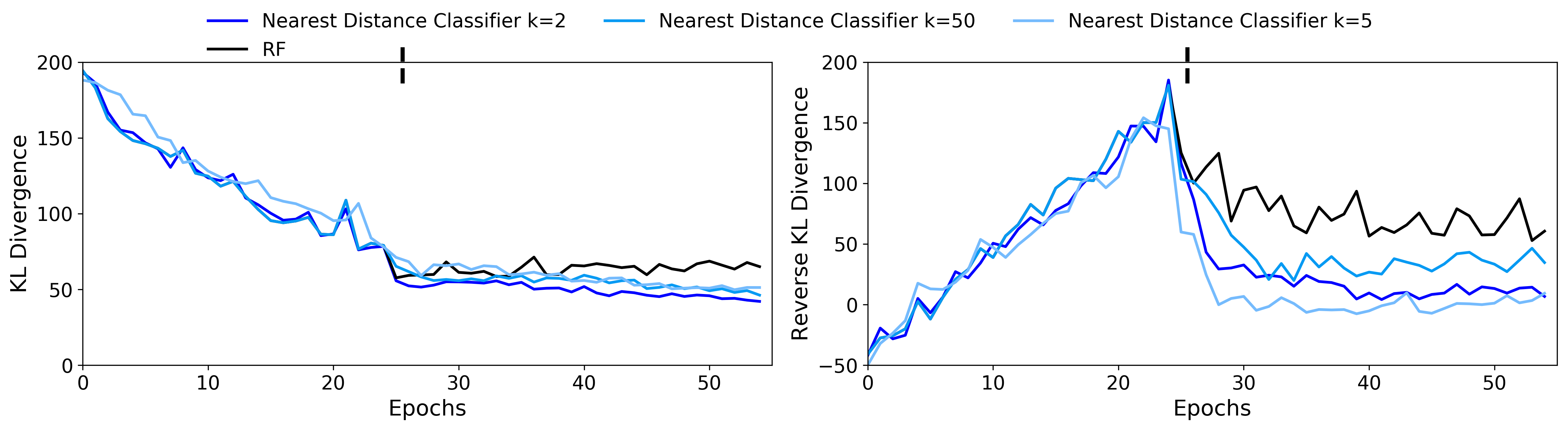}
\caption{{\bf Training progress on continuous MNIST with different critics.} RF = Random Forest.  
\label{fig:mnist-adversary-metrics}}
\end{figure*}

\begin{figure*}[t!]
\includegraphics[width=6.0in]{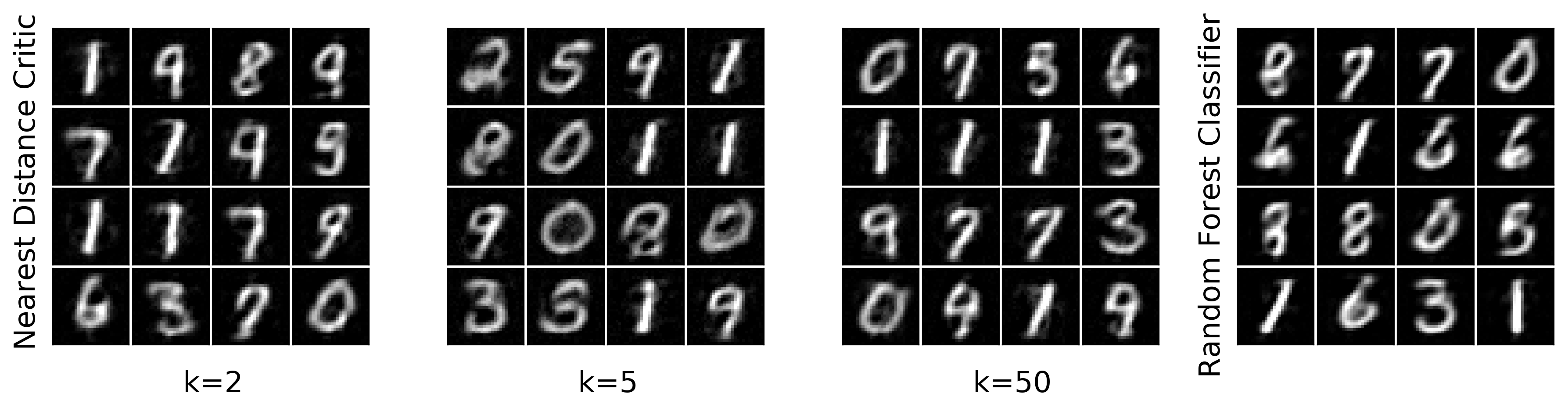}
\caption{{\bf Fantasy particles for continuous MNIST with different critics.} 
\label{fig:mnist-adversary-fantasy}}
\end{figure*}

\begin{figure*}[t!]
\includegraphics[width=3.0in]{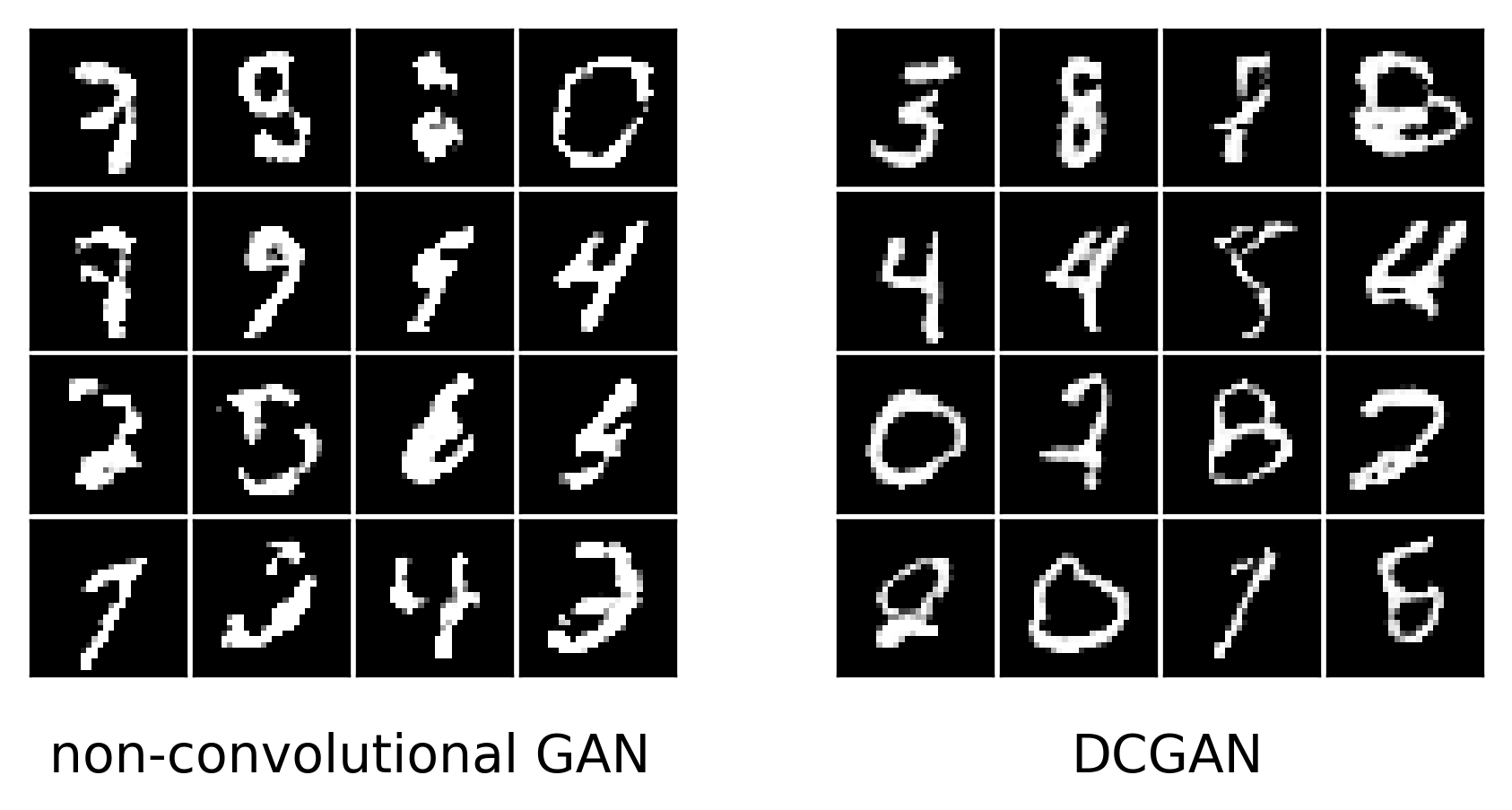}
\caption{{\bf Comparison of a non-convolutional GAN to a DCGAN on continuous MNIST.} 
\label{fig:dcgan-comparison}}
\end{figure*}

\subsubsection{Celebrity Faces}

\begin{figure*}[t!]
\includegraphics[width=6in]{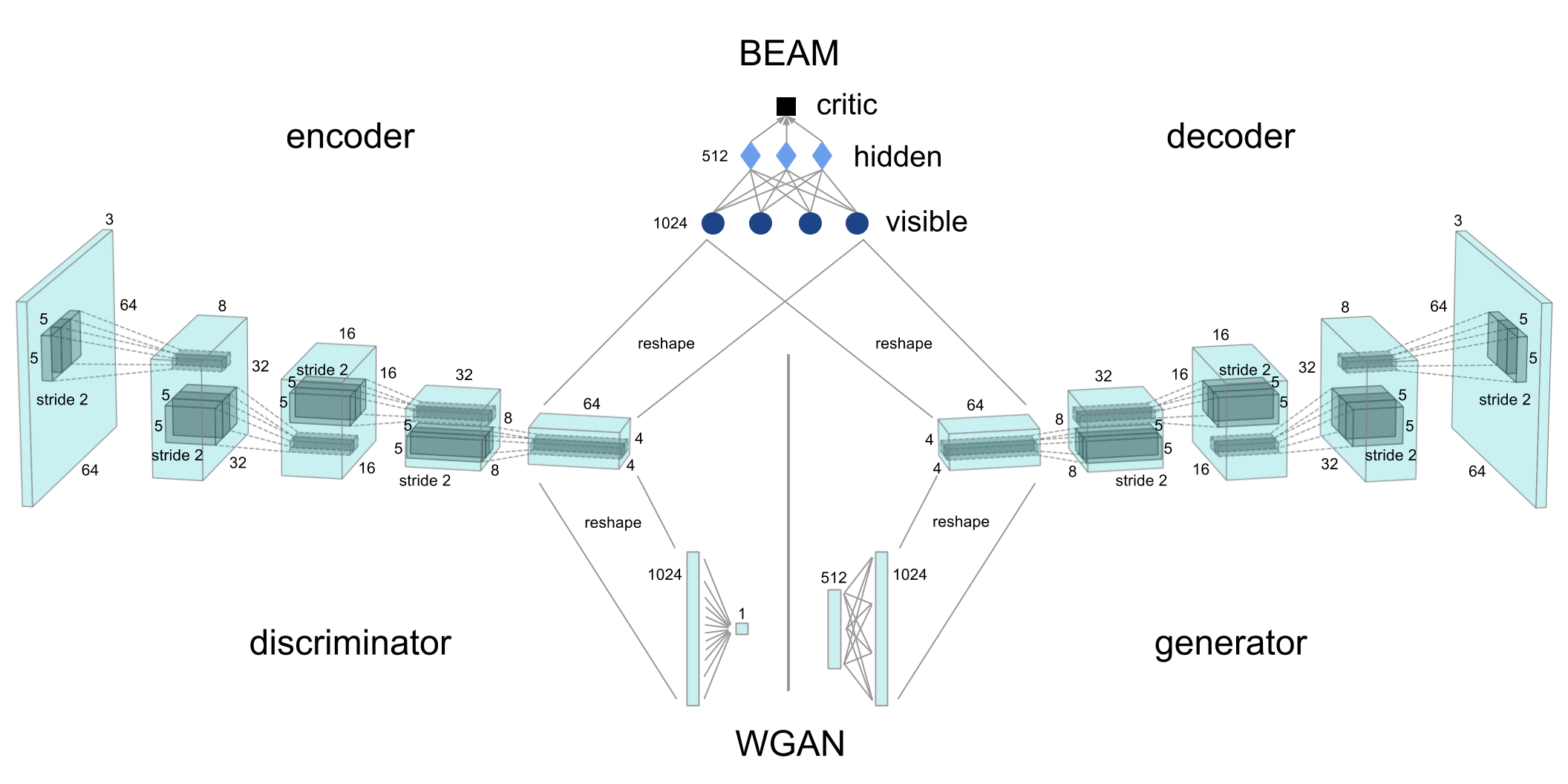}
\caption{{\bf Architectures used in the CelebA experiment.} The autoencoder is purely convolutional, with the encoder (decoder) shown by the stacks of convolutional (deconvolutional) layers.  The BEAM uses the flattened encoded features as the visible units.  The GAN/WGAN uses the same encoder (decoder) architectures for the discriminator (generator), with added single fully connected layers.  The autoencoder, BEAM, and GANs are each trained fully independently.  Table \ref{tab:celeb_params} lays out the parameters of the models.
\label{fig:celeb_arch}}
\end{figure*}

\begin{figure}[t]
\includegraphics[width=6.0in]{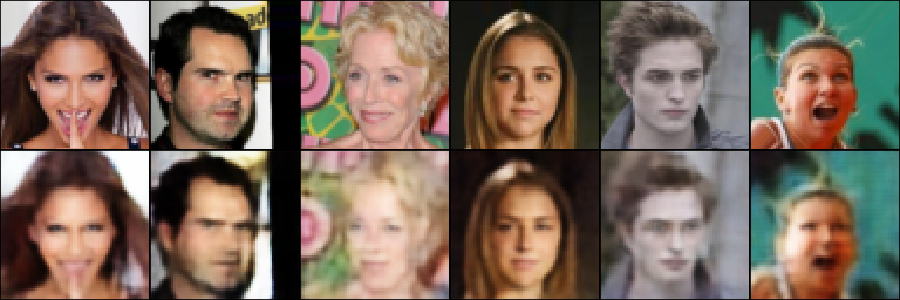}
\caption{{\bf Sample autoencoder reconstructions} The compression factor is $12288$ ($3 \times 64 \times 64$) to $1024$ ($64 \times 4 \times 4$).
\label{fig:celeb_recon}}
\end{figure}

\Fig{celeb_arch} shows a diagram of the architectures used in the CelebA dataset experiments for the BEAM (including the autoencoder) and GAN/WGAN.  The GANs share the same convolution architecture as the autoencoder, but are separately trained.

There is plenty of room to improve the quality of generated faces by employing more advanced RBM training techniques. For example, centering the RBMs tends to improve the variance in the generated faces and increases the definition of the hair and face edges.

\begin{figure*}[t]
\includegraphics[width=3.2in]{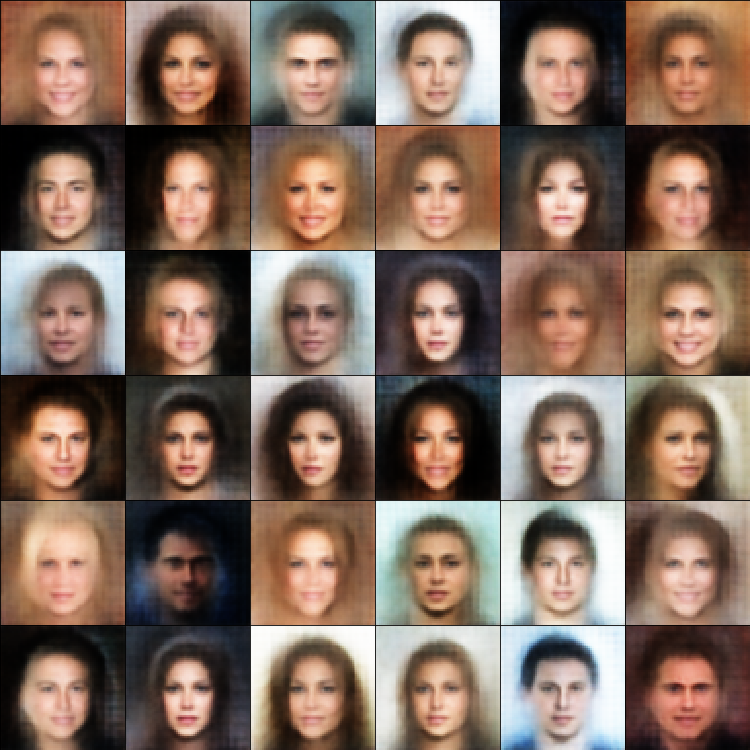}
\hspace{.1em}
\includegraphics[width=3.2in]{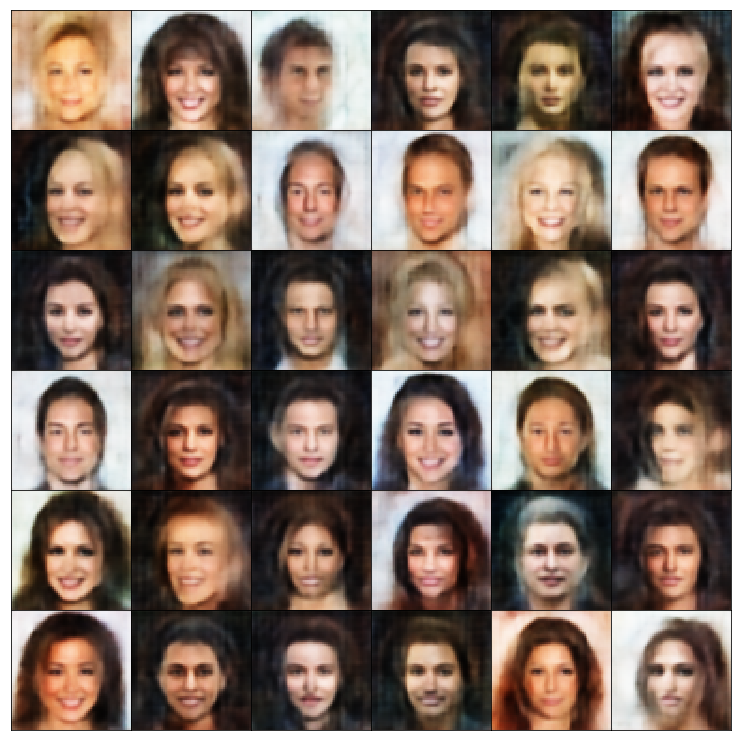}
\caption{{\bf BEAM vs. centered BEAM fantasy particles} Example fantasy particles generated by a BEAM using centered visible layer.
\label{fig:centered_beam}}
\end{figure*}
\begin{table}[h!]
\setlength\tabcolsep{2pt}
\begin{tabular}{|c|c|c|c|c|}
\hline
\bf{Bimodal Gaussian} & \multicolumn{4}{|c|}{$10^4$ samples, batch size $100$}\\
\hline\hline
GAN/WGAN & \multicolumn{4}{|c|}{fully-connected with ReLU activations, WGAN weight clamping $.1$} \\
\hline
\hphantom & generator dimensions & critic dimensions & epochs & lr \\
\hline
\hphantom & $1-32$ & $1-32-1$ & $100$ & $0.002$ \\
\hline
RBM/BEAM & \multicolumn{4}{|c|}{distance-weighted nearest-neighbor critic $k=5$, $\lambda=0.5$ for BEAM}\\
\hline
\hphantom & dims & MCMC steps & epochs & lr \\
\hline
\hphantom & $1-10$ & $100$ & $10$ & $.1$\\
\hline

\hline
\bf{Radial Gaussian} & \multicolumn{4}{|c|}{$10^4$ samples, batch size $100$}\\
\hline\hline
GAN/WGAN & \multicolumn{4}{|c|}{fully-connected with ReLU activations, WGAN weight clamping $.1$} \\
\hline
\hphantom & generator dimensions & critic dimensions & epochs & lr \\
\hline
\hphantom & $2-64-64$ & $2-64-64-1$ & $300$ & $0.001$ \\
\hline
RBM/BEAM & \multicolumn{4}{|c|}{distance-weighted nearest-neighbor critic $k=5$, $\lambda=0.5$ for BEAM}\\
\hline
\hphantom & dims & MCMC steps & epochs & lr \\
\hline
\hphantom & $1-10$ & $100$ & $30$ & $.15$\\
\hline
\hline
\bf{Grid Gaussian} & \multicolumn{4}{|c|}{$10^5$ samples, batch size $1000$}\\
\hline\hline
GAN/WGAN & \multicolumn{4}{|c|}{fully-connected with ReLU activations, WGAN weight clamping $.1$} \\
\hline
\hphantom & generator dimensions & critic dimensions & epochs & lr \\
\hline
\hphantom & $2-128-128-128$ & $2-128-128-128-1$ & $300$  & $.00001$ \\
\hline
RBM/BEAM & \multicolumn{4}{|c|}{distance-weighted nearest-neighbor critic $k=5$, $\lambda=0.5$ for BEAM}\\
\hline
\hphantom & dims & MCMC steps & epochs & lr \\
\hline
\hphantom & $1-20$ & $100$ & $200$ & $.2$\\
\hline
\end{tabular}
\caption{{\bf Gaussian mixture architectures and hyperparameters} All GAN/WGAN models use ReLU activations between fully-connected layers.  Network weights are initialized with normal distributions of standard deviation $0.2$, with biases zero-initialized.  The beta standard deviation for the driven sampler is set to $0$ for RBM, $.9$ for driven RBM and BEAMs.  The RBMs' learning rates decrease according to a power-law decay, and all training uses ADAM optimization with beta $= (0.5,0.9)$.
\label{tab:gmm_params_table}}
\end{table}

\begin{table}[h!]
\setlength\tabcolsep{2pt}
\begin{tabular}{|c|c|c|c|c|}
\hline
\bf{MNIST} & \multicolumn{4}{|c|}{$6 \cdot 10^5$ samples, batch size $100$}\\
\hline\hline
GAN/WGAN & \multicolumn{4}{|c|}{fully-connected with ReLU activations, sigmoid on generator, WGAN weight clamping $.01$} \\
\hline
\hphantom & generator dimensions & critic dimensions & epochs & lr \\
\hline
\hphantom & $100-164-164$ & $784-164-164-1$ & $55$  & $.001$ \\
\hline
RBM/BEAM & \multicolumn{4}{|c|}{distance-weighted nearest-neighbor critic $k=2$, $\lambda=0.1$ for BEAM}\\
\hline
\hphantom & dims & MCMC steps & epochs & lr \\
\hline
\hphantom & $784-200$ & $5$ & $25$ ML, $30$ adv. & $.001$ ML, $.0001$ adv.\\
\hline
\end{tabular}
\caption{{\bf Gaussian mixture architectures and hyperparameters} All GAN/WGAN models use ReLU activations between fully-connected layers.  Generator and discriminator weights are initialized with normal distributions of standard deviation $0.1$ and $0.02$ resp., with biases zero-initialized. All training uses ADAM optimization with beta $= (0.5,0.9)$ for the GANs and $(0.9,0.999)$ for the BEAM.  For the BEAM, the beta standard deviation for the driven sampler is set to $.95$. The RBMs' learning rates decrease according to a power-law decay.
\label{tab:mnist_params}}
\end{table}

\begin{table}[t]
{\tiny
\setlength\tabcolsep{2pt}
\begin{tabular}{|c|c|c|c|c|}
\hline
\bf{CelebA} & \multicolumn{4}{|c|}{$202599$ samples, batch size $128$}\\
\hline\hline
GAN/WGAN & \multicolumn{4}{|c|}{conv. with batch-norm and ReLU activations, WGAN weight clamping $.01$} \\
\hline
\hphantom & generator dimensions & critic dimensions & epochs & lr \\
\hline
\hphantom & \mbox{\tiny $64 \times 4^2-64 \times 8^2 64\times 16^2-32\times32^2-3\times 64^2$} & \mbox{\tiny $3\times 64^2-32\times32^2-64\times16^2-64\times8^2-64\times4^2$} & $10$  & $.0002$ \\
\hline
BEAM & \multicolumn{4}{|c|}{distance-weighted nearest-neighbor critic $k=5$, $\lambda=0.1$ for BEAM}\\
\hline
\hphantom & dims & MCMC steps & epochs & lr \\
\hline
\hphantom & $1024-512$ & $100$ & $10$ ML, $40$ adv. & $10^{-3}$ ML, $10^{-4}$ adv.\\
\hline
\end{tabular}
}
\caption{{\bf CelebA architectures and hyperparameters} All GAN/WGAN models use ReLU activations between fully-connected layers.  Generator and discriminator weights are initialized with normal distributions of standard deviation $0.02$, with biases zero-initialized. The beta standard deviation for the driven sampler is set to $0$ for RBM, $.95$ for the BEAM. All training uses ADAM optimization with beta $= (0.5,0.9)$ for the GANs and $(0.9,0.999)$ for the BEAM. The RBMs' learning rates decrease according to a power-law decay.
\label{tab:celeb_params}}
\end{table}